\documentclass[lettersize,journal]{IEEEtran}
\usepackage[pdftex]{graphicx}
\usepackage[caption=false,font=normalsize,labelfont=sf,textfont=sf]{subfig}

\usepackage{fixltx2e}
\usepackage{booktabs}
\usepackage{graphicx}
\usepackage{subcaption}
\usepackage{amsmath,amsfonts}
\usepackage{algorithmic}
\usepackage{algorithm}
\usepackage{array}
\usepackage[caption=false,font=normalsize,labelfont=sf,textfont=sf]{subfig}
\usepackage{textcomp}
\usepackage{stfloats}
\usepackage{url}
\usepackage{verbatim}
\usepackage{cite}
\usepackage{fancyhdr} 
\usepackage{xcolor}

\newcommand{\lnorm}[1]{\frac{#1}{\left\lVert{#1}\right\rVert _2}}

\bstctlcite{IEEEexample:BSTcontrol}
\title{Enhancing Content Representation for AR Image Quality Assessment Using Knowledge Distillation}

\author{Aymen Sekhri$^{1, 2}$, Seyed Ali Amirshahi$^{2}$, Mohamed-Chaker Larabi$^{1}$~\IEEEmembership{Senior Member,~IEEE} \\ 
        \textit{$^{1}$CNRS, Université de Poitiers, XLIM, Poitiers, France} \\ 
        \textit{$^{2}$Norwegian University of Science and Technology, Gjøvik, Norway} 
}

\begin{document}

\maketitle

\begin{abstract}
Augmented Reality (AR) is a major immersive media technology that enriches our perception of reality by overlaying digital content (the foreground) onto physical environments (the background). It has far-reaching applications, from entertainment and gaming to education, healthcare, and industrial training. Nevertheless, challenges such as visual confusion and classical distortions can result in user discomfort when using the technology. Evaluating AR quality of experience becomes essential to measure user satisfaction and engagement, facilitating the refinement necessary for creating immersive and robust experiences. Though, the scarcity of data and the distinctive characteristics of AR technology render the development of effective quality assessment metrics challenging. This paper presents a deep learning-based objective metric designed specifically for assessing image quality for AR scenarios. The approach entails four key steps, (1) fine-tuning a self-supervised pre-trained vision transformer to extract prominent features from reference images and distilling this knowledge to improve representations of distorted images, (2) quantifying distortions by computing shift representations, (3) employing cross-attention-based decoders to capture perceptual quality features, and (4) integrating regularization techniques and label smoothing to address the overfitting problem. To validate the proposed approach, we conduct extensive experiments on the ARIQA dataset. The results showcase the superior performance of our proposed approach across all model variants, namely TransformAR, TransformAR-KD, and TransformAR-KD+ in comparison to existing state-of-the-art methods.
\end{abstract}


\section{Introduction}

\IEEEPARstart{A}{ugmented} Reality (AR), along with other advanced immersive media technologies such as Virtual Reality (VR) and Mixed Reality (MR), represent the next frontier in display technologies \cite{cakmakci2006head}. Unlike traditional flat screens like those found on mobile phones or computers, these technologies aim to revolutionize how users interact in their surrounding environment \cite{zhan2020augmented}. AR, in particular, stands out as a groundbreaking innovation that enriches real-world environments by seamlessly integrating computer-generated information into users' surroundings through devices like smartphones, tablets, glasses, or Head-Mounted Displays (HMDs). This integration creates immersive and interactive experiences that extend beyond the confines of conventional reality. Applications of AR span various fields including navigation, education, entertainment, and healthcare \cite{vertucci2023history, 9897600}. However, ensuring the user satisfaction with AR experiences depends on meeting quality expectations. To accomplish this, it is crucial to grasp the concept of Quality of Experience (QoE) in AR and to understand the fundamental characteristics of immersive media. The latter refers to a psychological state where individuals feel surrounded, included, and engaged within an environment that continuously delivers stimuli and experiences \cite{itu_g1035}. To do so, immersive systems may use technologies such as displays (e.g. VR, AR, 4k HDR, 8k, etc.), accurate positional tracking, and haptic feedback \cite{perkis2020qualinet}. In another work \cite{van2016contextual}, immersion is created through six dimensions,  namely presence, perspective, proximity, point of view, participation, and place.

To develop objective metrics for AR Image Quality Assessment (AR-IQA) or the QoE in immersive media in general, a profound understanding of the Human Visual System (HVS) \cite{mazin2023research, hodvzic2018dashrestreamer, dziembowski2022iv, duanmu2023bayesian} and the various influencing factors \cite{itu-rec-g1036} are essential. Traditionally, the gold standard for measuring QoE involves conducting psychophysical (subjective) experiments, where the Mean Opinion Score (MOS) is used as the ground truth for the assessment. However, for AR, this process is complex, time-intensive, and requires meticulous design and planning to ensure a realistic scenarios. These challenges, in addition to the recency of AR, contribute to the scarcity of subjective data for AR-IQA. Despite these challenges, various studies have focused on Subjective Quality Assessment (SQA) for other types of immersive media, providing valuable insight that can be leveraged for AR-IQA. 

Starting with 360-degree images, Sendjasni et al. \cite{sendjasni2023objective} focused on the effects of using multiple HMD manufacturers. The study included eight valid observers and used Varjo VR-2, HTC Vive Pro, HP Reverb VR, and Oculus Quest. They created the 360-IQA database, which contains 240 distorted versions of 20 pristine 360-degree images, applying different levels of distortions such as JPEG compression, blur, and white Gaussian noise. Duan et al. \cite{duan2023attentive} investigated the influence of distortions caused by the omnidirectional stitching process of the dual fisheye images in 360-degree images using HTC VIVE Pro. Enhancing realism, researchers are shifting their focus to omnidirectional videos and incorporating audio signals. Zhu et al. \cite{zhu2023perceptual} for example, proposed a database with 360 omnidirectional audio-visual content, arguing that the audio signal plays a crucial role in the QoE. Another work focused on omnidirectional videos is done by Elwardy et al. \cite{elwardy2023acr360} in which they propose a dataset of 360-degree videos that provide psychophysical and psychophysiological data by running a subjective test using a headset under two different rating strategies.
As one of the crucial aspects that affect the QoE in immersive media is providing depth information, therefore many studies have focused on working on stereoscopic images. Xu et al. \cite{xu2018subjective} created a stereoscopic omnidirectional IQA database to evaluate the overall quality of these images in a VR environment. The experiments were conducted using Samsung Gear VR under various depth levels, achieved by altering the disparity level between the two views presented to the subject from zero disparity to medium and large. Additionally, they investigated quality variations by employing different quantization parameters for a better portable graphics format. It is worth noting that the reference in this dataset is distorted either symmetrically or asymmetrically. Chen et al. \cite{chen2019study} create a diverse and representative immersive stereoscopic 3D image quality database to study and model the effect of different types of VR-related distortions like Gaussian noise, blur, stitching distortion, downsampling distortion, VP9 compression, and H.265 compression on this type of media. Another work presented by Escobar et al. \cite{moreno20223d22} creates a dataset of stereoscopic images in diverse indoor and outdoor settings and orchestrates a comprehensive psychological experiment to assess the overall QoE. Additionally, they investigate potential discomfort effects on participants to gain deeper insights into the nuanced responses and reactions to stereoscopic imagery. 

SQA for AR has received limited attention in the research landscape. Most existing studies have predominantly addressed geometric degradations affecting digital objects in AR, including point clouds or 3D meshes, often against a blank or uniform textured background \cite{9502695}. Guo et al. \cite{guo2016subjective} evaluate the visual quality of the 3D meshes under different type of distortions by showing the participants a low-speed rotation animation around the vertical axis, with a blank background. Gutiérrez et al. \cite{gutierrez2020quality} assess the quality of the 3D meshes using an HMD instead of 2D displays under different lighting conditions and provide recommendations for subjective testing of QoE in MR/AR scenarios using optical see-through HMD, namely Microsoft HoloLens. In 2017, Alexiou et al. \cite{alexiou2017towards} studied the effect of the geometric and texture degradations that affect point clouds using an HMD. 

\begin{figure}[t!]
    \centering
    \includegraphics[width=1.0\columnwidth]{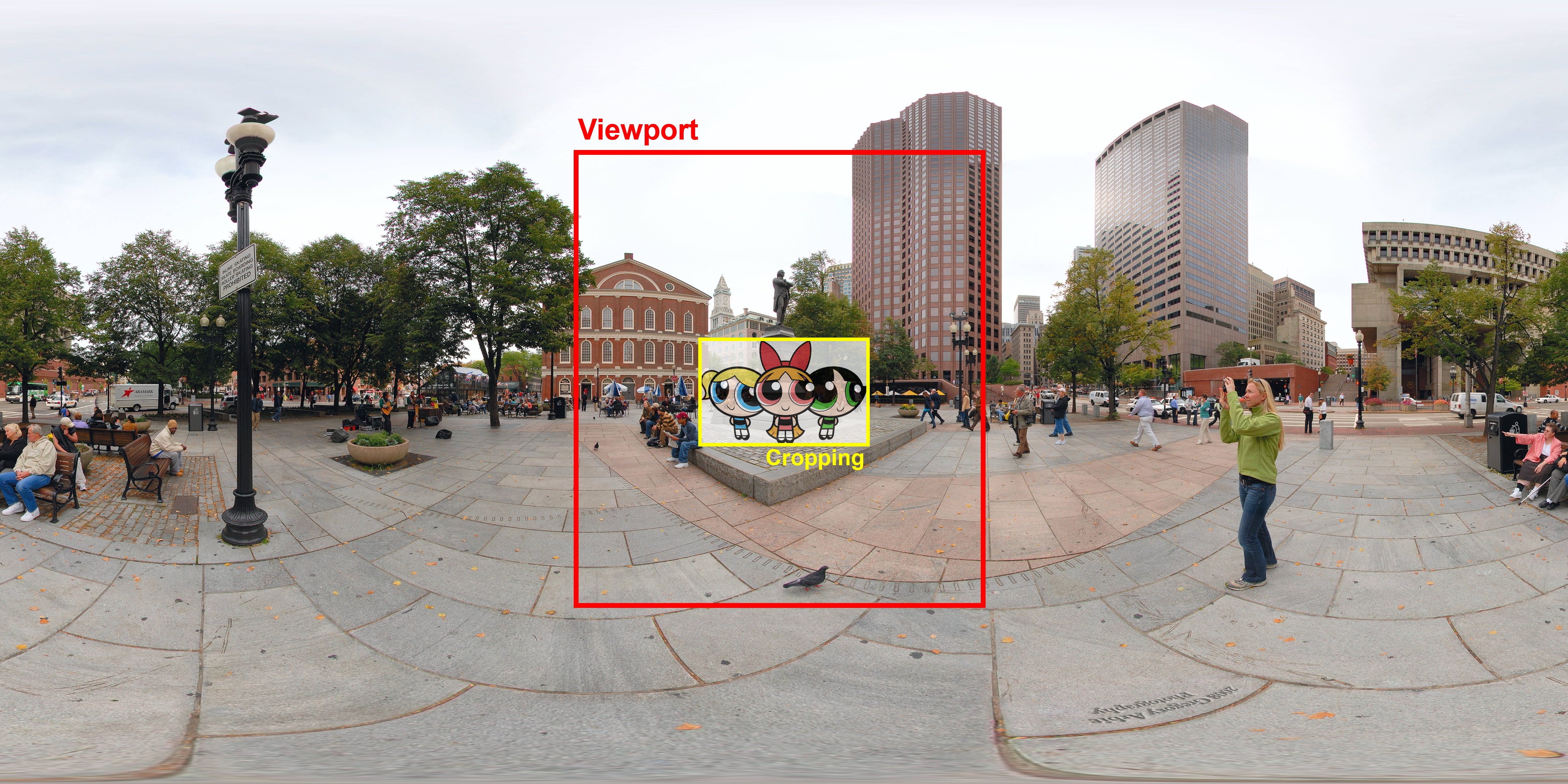}
    \caption{AR image (foreground) superimposed on the background image and the viewport captured during the subjective test, and the cropped region is used for the assessment of the objective metric.}
    \label{fig:viewport}
\end{figure}

A recent study by Duan et al. \cite{duan2022confusing} has taken a comprehensive approach to SQA in the field of AR. Unlike conventional studies, this research expands its scope to encompass the concept of visual confusion between digital objects and the real-world background. In the work, Duan et al. \cite{duan2022confusing} assessed the impact of monocular visual confusion in AR scenarios using the HTC Vive Pro Eye. The study included twenty-three subjects and evaluated 560 experimental stimuli by superimposing digital images onto omnidirectional images and including different types of distortions such as JPEG compression, image scaling, and image contrast adjustment, and they consider monocular visual confusion as the main distortion. Their research involves the development of two key datasets. The first, CFIQA (Confusing Image Quality Assessment), offers insights into human perception of superimposed images with varying mixing thresholds, while the second, ARIQA dataset, simulates a more realistic AR application scenarios by overlaying three types of images namely web, natural, and graphical onto omnidirectional images to mimic real-world backgrounds (Figure \ref{fig:viewport}).

In the realm of Full-Reference Image Quality Assessment (FR-IQA) where we have access to both the reference and the test images, numerous approaches have emerged for evaluating image quality. These methods vary in their strategies for extracting features and patterns that maximize the correlation between predicted quality scores and the subjective scores. They can be broadly categorized into classical methods, which involve a meticulous engineering process to extract prominent features, and learnable IQA methods, where features are learned by training deep learning-based models on a SQA dataset.

Classical metrics such as Mean Squared Error (MSE) measures the distortion between the reference and the distorted image by calculating the average squared difference of pixel values. It measures the discrepancy between the original and distorted images on a pixel-by-pixel basis. Peak Signal-to-Noise Ratio (PSNR), an extension of MSE, quantifies the fidelity of the image reconstruction by measuring the ratio between the maximum possible power of a signal and the power of noise, providing a quantitative assessment of image fidelity. Higher PSNR values indicate better image quality. Another notable metric is the Structural SIMilarity index (SSIM), proposed by Wang et al. \cite{wang2004image}, which evaluates the similarity in local patches by analyzing three main components; luminance, contrast, and structure. On one hand, these components collectively contribute to the overall metric. The latter provides a more holistic evaluation of image similarity beyond pixel-level differences, however multiple variants of SSIM have been developed \cite{wang2003multiscale, wang2010information}. On the other hand, the Feature SIMilarity index (FSIM), proposed by Zhang et al. \cite{zhang2011fsim}, diverges from the SSIM approach by prioritizing the comparison of structural features within images. FSIM computes similarity based on the resemblance of feature maps extracted from the reference and distorted images. This method aims to capture higher-level visual information beyond mere pixel-level comparisons, thereby offering a more comprehensive measure of image similarity. Despite the effectiveness of classical methods, they often struggle to correlate well with the HVS, particularly in scenarios involving immersive media, such as AR and/or VR where extracting prominent quality features is crucial. Moreover, these methods frequently introduce more intricate hand-crafted features to quantify image dissimilarities.

Learnable approaches have been proposed recently to overcome the limitations of hand-crafted features. Kang et al. \cite{kang2015simultaneous} introduced an approach for simultaneously estimating image quality and identifying distortions. Their work presented compact multi-task Convolutional Neural Networks (CNNs), leveraging the inherent advantages of CNNs for addressing multi-task problems. Some studies use convolutional sparse coding for distortion identification and subsequent quality assessment \cite{yuan2015image}. Bosse et al. \cite{bosse2017deep} introduced FR-IQA and no reference IQA metrics by jointly learning local quality and local weights within a unified framework, without relying on hand-crafted features or prior domain knowledge. One widely used approach in IQA is the Learned Perceptual Image Patch Similarity (LPIPS) \cite{zhang2018unreasonable}, which is considered a well-established framework for perceptual judgment tasks. This framework involves computing distance features across the channel dimension of the feature maps produced by CNN-based feature extractors, specifically SqueezeNet \cite{iandola2016squeezenet}, AlexNet \cite{krizhevsky2012imagenet}, and VGG \cite{simonyan2014very}. Subsequently, averaging these distance features across the spatial dimension yields distance information. The latter is then mapped to predict perceptual judgment, such as predicting the MOS. Notably, given the success of the transformer architecture across various tasks, researchers have begun exploring its use in IQA tasks by encoding the input image with CNNs and then feeding the features to an encoder-decoder transformer to model quality features and overcome the limited receptive filed of CNNs \cite{cheon2021perceptual, golestaneh2022no, jiang2022image}.

A reliable Objective Quality Assessment (OQA) model does not only serves as an image quality monitor but also plays important roles in optimizing various quality-driven applications, such as image/video coding \cite{wang2012perceptual}, image fusion \cite{ma2015perceptual}, contrast enhancement \cite{gu2014automatic}, and so on. At present, AR-IQA metrics are largely lacking, primarily due to the scarcity of data and the inherent complexity of such technologies. Despite these challenges, Duan et al.  \cite{duan2022confusing} evaluate AR image quality using classical FR-IQA metrics on the ARIQA dataset. The metrics tested include PSNR, NQM \cite{damera2000image}, SSIM \cite{wang2004image}, MS-SSIM \cite{wang2003multiscale}, VIF \cite{sheikh2006image}, IW-MSE, IW-PSNR, IW-SSIM \cite{wang2010information}, FSIM \cite{zhang2011fsim}, GSI \cite{liu2011image}, GMSD \cite{xue2013gradient}, GMSM \cite{xue2013gradient}, PAMSE \cite{xue2013perceptual}, LTG \cite{gu2014efficient}, and VSI \cite{zhang2014vsi}. These metrics have demonstrated limitations in AR-IQA, underscoring the necessity for more advanced IQA metrics to measure the perceptual quality of such images. Consequently, they investigate the LPIPS \cite{zhang2018unreasonable}. Finally, they proposed a CNN-based metric named, CFIQA model which uses VGG \cite{simonyan2014very} or ResNet-50 \cite{he2016deep} as feature extractors. These models produce features at each convolution layer from both reference and superimposed images. These features contain low-level characteristics due to the behavior of hierarchical feature maps produced by CNNs. Feature distance vectors are then generated by subtracting and normalizing the superimposed features from both reference images. This step is followed by channel attention to produce distance maps. The latter are then multiplied by spatial attention maps from a saliency prediction module, capturing high-level features, which allow predicting distance scores at each layer. The final score is the average of the distance scores. The framework is extended to AR with the ARIQA model. The only difference is the use of two superimposed images that come from the same reference images, yet with different quality scores. An enhanced version of the ARIQA model, referred to as ARIQA+, incorporates features from the edge detection model. The performances of the proposed metrics are acceptable but they are still far below those for 2D quality.

The field of AR quality assessment is still emerging, requiring further development to achieve reliable quality predictions. Currently, many deep learning models are trained on natural images, making it difficult to adapt these models to effectively capture high-level semantic information and accurately assess perceptual quality in the presence of the visual confusion in AR images \cite{duan2022confusing}. Additionally, replicating the HVS in quality assessment is essential. This process involves scanning the entire image and focusing on specific areas to judge its quality \cite{cheon2021perceptual}. Human perception may also be influenced by the nature of the image content \cite{van2012influence}. Developing a learnable method to address these challenges may require complex models. Therefore, to tackle these challenges, we propose a novel FR-IQA method based on a lightweight transformer architecture and Knowledge Distillation (KD). The contributions of our work are summarized below:

\begin{itemize}
    \item A lightweight transformer-based framework for AR-IQA, containing four stages; content-aware encoders that capture long-range information based on the Vision Transformer (ViT) architecture with trained weights in a Self-Supervised Learning (SSL) manner. Using ViTs assissts in overcomming the limited receptive field associated with CNNs \cite{raghu2021vision}. The framework addresses shift representation caused by distortion and visual confusion, incorporates quality-aware decoders for modeling quality features, and includes final regressors for prediction.
    \item Using knowledge distillation to enhance learned abstract representations of the content-aware encoders in the presence of visual confusion. Our hypothesis suggests that providing the encoders with explicit foreground and background category information will enhance representation learning ability, resulting in the generation of richer quality features. 
    \item Employing label smoothing for IQA tasks to reduce model overconfidence in predicting the MOS in the training set.
    \item Our method outperforms state-of-the-art approaches on the ARIQA dataset.
\end{itemize}

The structure of the remainder of the paper is as follows, Section \ref{sec:qa_meth} elaborates on each component of our proposed quality assessment approach. Section \ref{sec:train_impl} provides details regarding the training implementation and the ARIQA dataset used in our study. A quantitative analysis, including comparisons with state-of-the-art methods, and an ablation study highlighting the significance of each component is given in Sections \ref{sec:res_disc} and \ref{sec:qual_analysis}. Additionally, qualitative analysis is conducted in these sections through visualization of attention maps and the use of Uniform Manifold Approximation and Projection (UMAP) \cite{mcinnes2018umap} to visualize the learned features. The paper concludes in Section \ref{sec:conclusion} with final remarks and outlines potential avenues for future research.

\section{Quality Assessment Method}
\label{sec:qa_meth}
Our foundational approach, named TransformAR, builds upon the transformer architecture \cite{vaswani2017attention} to tackle the task of quality assessment. To ensure clarity and depth of comprehension, we delineate the key components and methodology, starting with the content-aware encoders, proceeding to the quality-aware decoders, and concluding with the regressors. Subsequently, we explore the specifics of our proposed method, demonstrating the learning strategy and how enhancing the representation learning capabilities using knowledge distillation contributes to overall improvement.

\begin{figure}[t!]
    \centering
    \includegraphics[width=0.8\columnwidth]{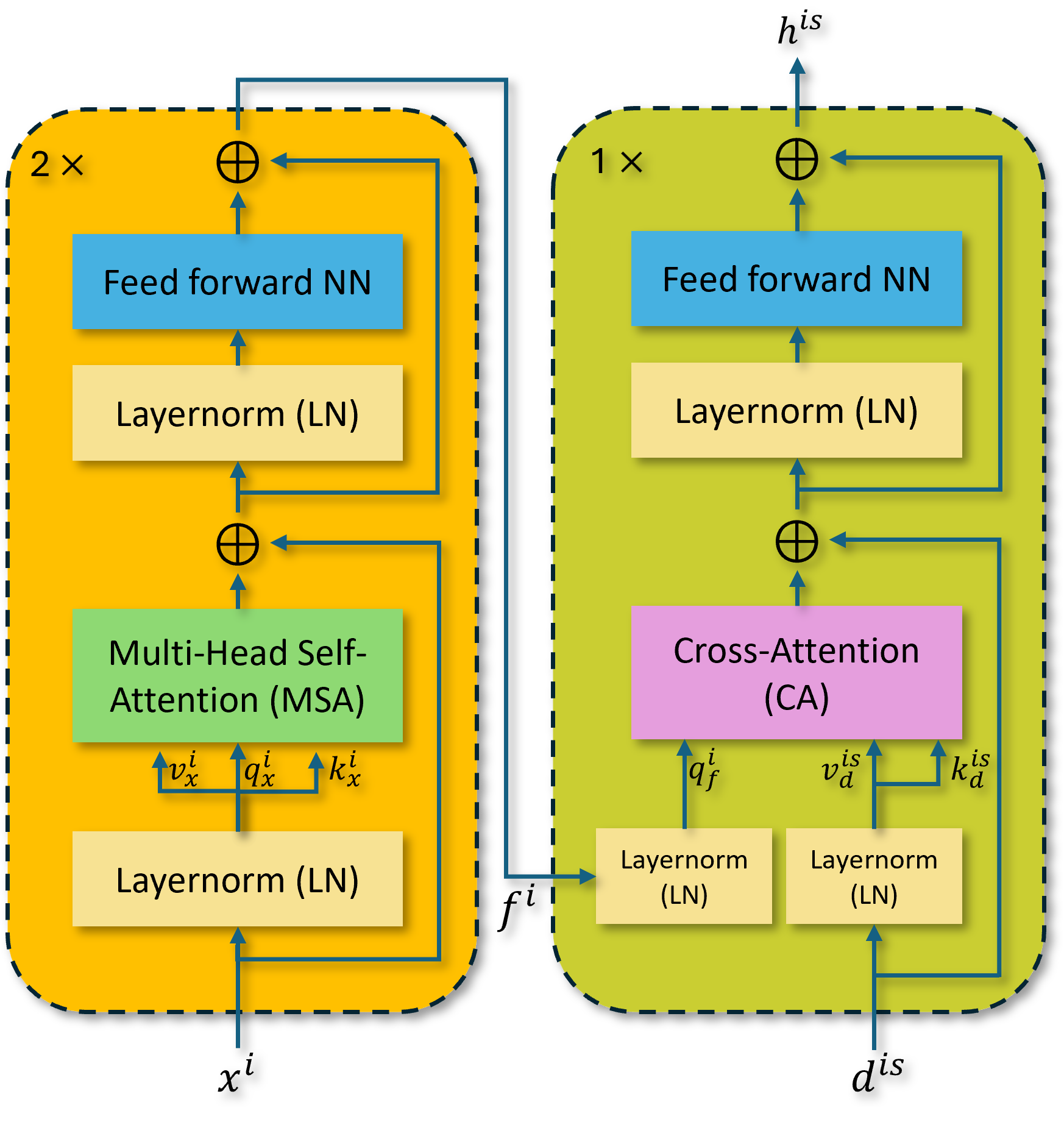}
    \caption{Illustration showcasing the encoder-decoder transformer architecture. (left) Two transformer encoder blocks are stacked, each preceded by Layernorm and followed by MSA. In addition, feed-forward neural networks and skip connections are used. (right) The decoder consists of a single transformer decoder block, preceded by Layernorm and followed by cross-attention mechanism, along with feed-forward neural networks and skip connections.}
    \label{fig:enc-dec}
\end{figure}

\subsection{Content-aware encoders}
\label{ssec:cont_enc}
The content-aware encoders are designed to map the input images into abstract representations that encapsulate semantic information. These encoders are based on a tiny ViT architecture. Following this encoding process, the shift representation induced by distortions is computed between the reference representations and the superimposed (distorted) one. We begin by introducing the ViT framework and then proceed to discuss why we used pretrained weights from a ViT model trained using self-supervised learning. Finally, we explain the calculation of the shift representation.

\subsubsection{Vision Transformer}
\label{sssec:vit}
ViT, is an architecture introduced by Dosovitskiy et al. \cite{dosovitskiy2020image}, representing a paradigm shift in computer vision by leveraging the  capabilities of transformer model's, first introduced in NLP \cite{vaswani2017attention}. Unlike CNNs that rely on local convolution filters for feature extraction, ViT adopts a self-attention mechanism to capture long-range dependencies among image patches. As transformers are built primarily to handle sequences of 1D vectors, the ViT model takes a 2D image $I \in \mathbb{R}^{H \times W \times 3}$ as input, where $H$ and $W$ represent the image height and width, respectively. The image is divided into $N$ patches, each of size $P \times P \times 3$. These patches are then flattened into a sequence of 1D vectors, called tokens which are projected through a linear layer, resulting in a sequence of embedding tokens $x_p \in \mathbb{R}^{N \times C}$, where $C$ is the embedding dimension and $N$ is the number of tokens in the sequence. Additionally, a special learnable token $x_{\text{class}}$, is inserted into the sequence $x_p$, resulting in $N+1$ input embeddings. This token acts as the representation of the entire image. Subsequently, Position Embeddings (PE) are added to preserve positional information. The resulting sequence of embedding tokens serves as input to the transformer encoder.

The transformer encoder, as used in \cite{vaswani2017attention}, consists of $L$ transformer layers, each containing four main operations. First, a Layer Normalization (LN) is applied before every transformer layer, followed by the main modules, which are the Multi-headed Self-Attention (MSA), the Feed-Forward Neural Networks (FFNN), and the residual connections (Figure \ref{fig:enc-dec}). These modules are the building blocks of a transformer layer. Formally, the output of the $l$-th layer $H_l$ can be expressed as
\begin{equation}
    \left\{
        \begin{aligned}
            H_{l}' &= \text{MSA}(\text{LN}(H_{l-1})) + H_{l-1} \\
            H_l &= \text{FFNN}(\text{LN}(H_{l}')) + H_{l}'
        \end{aligned}
    \right. \text{where, } l \in [1, L].
\end{equation}
Thus, when $l = 1$, $H_{0}$ denotes the input embeddings. The core operation in the MSA is the self-attention mechanism. Given a sequence of embedding tokens, a set of queries $q = \{q_i\}_{i=0}^{N}$, keys $k = \{k_i\}_{i=0}^{N}$, and values $v = \{v_i\}_{i=0}^{N}$ are produced using learnable matrices. The self-attention mechanism computes the dot products of the query with all keys, divides each by $d_k$ which represents the dimension of the key vectors, and applies a softmax function to obtain the weights on the values. The attention formula is as follows:
\begin{equation}
    \text{Attention}(q, k, v) = \textit{softmax}\left(\frac{qk^T}{\sqrt{d_k}}\right)v
\end{equation}

ViT offers several advantages over traditional CNNs, including enhanced scalability and ability to handle input images of different resolutions. Moreover, by leveraging the self-attention mechanism, ViT captures long-range dependencies among image patches, facilitating better contextual understanding and going beyond the limited receptive field that CNNs suffer from \cite{raghu2021vision}. Dosovitskiy et al. \cite{dosovitskiy2020image} demonstrated that ViT achieves competitive performance on various vision tasks, including image classification, object detection, and semantic segmentation, surpassing or matching the performance of state-of-the-art CNN architectures.

\subsubsection{Self-supervised Learning}
\label{sssec:ssl}
SSL is a machine learning paradigm wherein a model learns representations from input data without explicit supervision labels \cite{shwartz2023compress}. It exploits inherent data structures and characteristics. SSL offers numerous advantages, including leveraging unlabeled data, enhancing generalization by observing various data aspects, and reducing reliance on human annotation. There are different types of SSL \cite{balestriero2023cookbook}, encompassing the deep metric learning \cite{chen2020simple}, the knowledge distillation \cite{caron2021emerging}, and the canonical correlation analysis \cite{zbontar2021barlow, bardes2021vicreg}. These methods, among many others, employ either CNNs or ViTs as feature extractors. In our study, we use the self-DIstillation with NO labels (DINO)'s pretrained ViT \cite{caron2021emerging}. The latter's primary objective is to employ knowledge distillation to generate similar representations from different distorted views derived from the same image. It means that regardless of the transformations and applied distortions, the DINO encoder prioritizes extracting relevant content representations containing explicit semantic segmentation information \cite{caron2021emerging}.

\begin{figure*}[t]
  \centering
  \includegraphics[width=\textwidth]{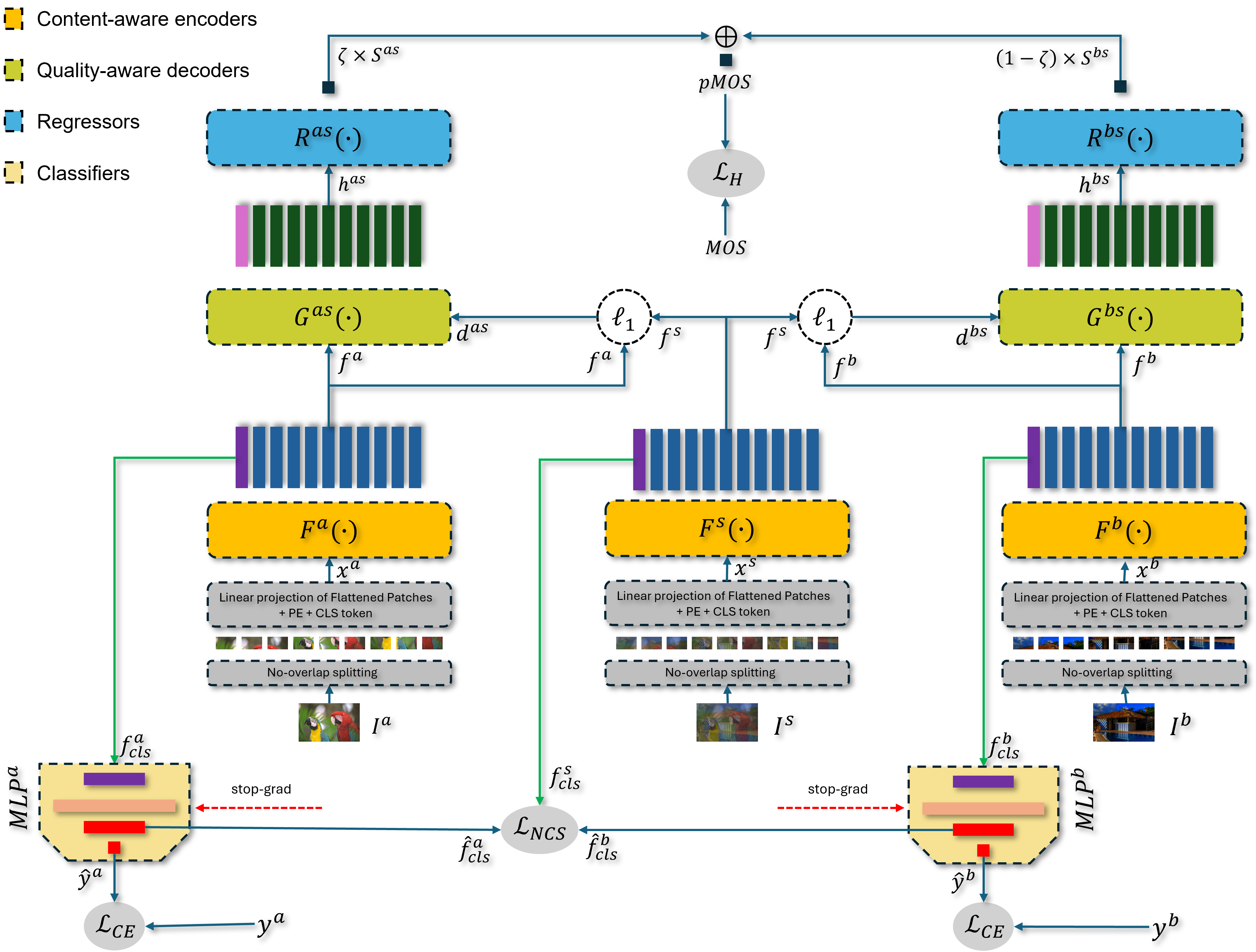} 
 \caption{Overview of the TransformAR-KD+ model. Content-aware encoders $\mathcal{F}^i(\cdot)$ extract abstract representations $f^i \in \mathbb{R}^{N\times C}$ from each input image $I^i$, where $i = \{a, b, s\}$. $MLP^a$ and $MLP^b$ are used to generate ground-truth representations $\hat{f^a_{cls}}$ and $\hat{f^b_{cls}}$, predicting the classes of references. Subsequently, the cosine similarity between $f^s_{cls}$ and the ground-truth representations is maximized, and the $l_1$-distance is employed to compute the shift caused by distortions. Quality-aware decoders $G^{as}(\cdot)$ and $G^{bs}(\cdot)$ align the information in the reference representations with relevant information in the shift representations using the cross-attention mechanism, while regressors $R^{as}(\cdot)$ and $R^{bs}(\cdot)$ map the quality representations to quality scores. These scores are aggregated to produce $pMOS$.}
  \label{fig:arch}
\end{figure*}

Various ViT architectures have been developed, often differing in the number of transformer layers $L$. We use the ViT-S/16 version of DINO, which represents the smallest variant with a patch size of $16 \times 16$ pixels. Hence, we have three input images, the background image $I^b$ which reflects the real-world, the AR image $I^a$, considered as the foreground, and finally the superimposed image $I^s$, which is the perceptually distorted image 
\begin{equation} \label{eq:mixing}
I^s = \sigma \circ D(I^a) + (1-\sigma) \circ I^b.
\end{equation}
In Equation \eqref{eq:mixing} $D(\cdot)$ denotes the applied distortions, and $\sigma$ is the mixing value. This results to three ViT encoders $\mathcal{F}^*(\cdot)$ (Figure \ref{fig:arch}). Consequently, for each input image, a sequence of embedding vectors $f^* = \{f^*_i\}_{i=0}^{N}$ is produced where $f^* = \mathcal{F}(I^*)$, and  $* = \{a, b, s\}$. However, we have noticed that our framework is prone to overfitting when using all twelve transformer layers in the ViT-S/16. Therefore, we empirically chose only the first two layers, which makes our framework extremely lightweight.

\subsubsection{Shift Representation}
The next step involves using the $l_1$ distance to compute the sequence of shift representation between superimposed (distorted) and reference embedding vectors. This maintains the true residual caused by distortions at each patch. $l_1$ distance was preferred over $l_2$ because the latter results in very small shift representations when the residuals are smaller than one, not accurately reflecting the actual shift caused by distortions. Shift representations are calculated by
\begin{align} \label{eq:l1}
    \left\{
    \begin{array}{l}
        d^{as}_{i} = |f^s_{i} - f^a_{i}| \\
        d^{bs}_{i} = |f^s_{i} - f^b_{i}| \\
    \end{array}
    \right.
    \text{Where, } i = 0, 1, \ldots, N
\end{align} 
where $d^{as}$ (resp. $d^{bs}$) denotes the shift between the output vectors of the AR image $I^a$ (resp. $I^b$) and the superimposed image $I^s$. Here, $d^{as}$ (resp. $d^{bs}$) and $f^a$ (resp. $f^b$) serve as input sequences to the quality-aware decoders $\mathcal{G}^{as}(\cdot)$ (resp. $\mathcal{G}^{bs}(\cdot)$), which will be described in Section \ref{ssec:quality_decoders}.

\subsection{Enhanced Representation Learning}
\label{ssec:enhc_rep}
By freezing the pretrained weights of the encoders $\mathcal{F}^i(\cdot)$, abstract representations of the input images are generated. However, unlike human perception, these representations may lack specific information about the characteristics of the AR image or the nature of the background. Instead, we opt to fine-tune the encoders. We incorporate an MLP projection head on top of the reference encoders $\mathcal{F}^a(\cdot)$ and $\mathcal{F}^b(\cdot)$, referred to as $MLP^a$ and $MLP^b$ respectively (Figure \ref{fig:arch}). The latter consists of three linear layers, an input layer mapping the output classification token, $f^{i}_{cls}$ with $i \in \{a, b\}$ to a higher latent space, a hidden layer bringing back the output to the original dimension with additional information about the class in the input image, resulting in $\hat{f}^{a}_{cls}$ (resp. $\hat{f}^{b}_{cls}$). The outputs will be used as ground-truth representations of the reference images. This will allow us to distill knowledge from both $\mathcal{F}^a \circ MLP^a$ and $\mathcal{F}^b \circ MLP^b$ (referred to as teacher encoders) to the superimposed encoder $\mathcal{F}^s$ (referred to as the student encoder). This process helps the encoder in disentangling information about both the foreground and background, enhancing its representation capability by incorporating details about the input AR image's category and the characteristics of its background, even in the presence of visual confusion.

The outputs $\hat{f}^{a}_{cls}$ and $\hat{f}^{b}_{cls}$ are projected to logits $\hat{z}^{a}$ and $\hat{z}^{b}$, respectively, with dimensions $C_{a}$ and $C_{b}$ representing the number of classes. For the AR reference image, which comprises of three classes (web, natural, and graphics), and the background reference image (outdoor and indoor classes), we apply a softmax activation function to obtain a probability distribution. The highest probability among these distributions corresponds to the predicted class. The equations for computing the logits and target vectors are given by:

\begin{align} 
    \begin{bmatrix} f^{i}_{cls} \\ f^{i}_{j} \end{bmatrix} &= \mathcal{F}^{i}\left( x^{i}_{cls} \oplus x^{i}_{j} \right) \\
    \begin{bmatrix} \hat{z}^{i} \\ \hat{f}^{i}_{cls} \end{bmatrix} &= MLP^{i}\left( f^{i}_{cls} \right)
\end{align}
where $i = a, b$ and $j = 1, 2, \ldots, N$. The MLPs and the reference encoders are updated in a supervised manner using ground-truth labels $y^{i}$ representing the class index, then minimizing the Cross Entropy (CE) loss defined as
\begin{align}
    \text{CE}(y, z) &= -\sum_{m}^{M} y_m \log(\textit{softmax}(z_m)).
\end{align}
Here, $y$ is the ground-truth labels and $z$ is the output logits, $M$ is the number of images within a batch, and the softmax is defined as

\begin{equation}
\textit{softmax}(x_t) = \frac{e^{x_t}}{\sum_{k=1}^{K} e^{x_k}}
\end{equation}
where $x_t$ is the input logits of the $t$-th predicted class, and $K$ is the number of classes. The overall classification loss function of the reference images is defined as:

\begin{equation}
\mathcal{L}_{CE} = \frac{1}{2}\text{CE}(y^{a}, \hat{z}^{a}) + \frac{1}{2}\text{CE}(y^{b}, \hat{z}^{b})
\end{equation}

The $x_{\text{class}}$ token within the superimposed image needs to encapsulate information regarding both the background and the AR image content  simultaneously, as the superimposed image comprises a blend of the two with a threshold $ \sigma $ indicated in Equation \eqref{eq:mixing}. To tackle this challenge, we choose to align the representation $ f^s_{cls} $ of the superimposed image with the ground-truth representations $ \hat{f}^{a}_{cls} $ (resp. $ \hat{f}^{b}_{cls} $). This alignment is accomplished by minimizing a specialized loss between $ f^{s}_{cls} $, which captures semantic details about the superimposed image, and $ \hat{f}^{a}_{cls} $ (resp. $ \hat{f}^{b}_{cls} $), which contain sufficient information about the content present in the reference images. The equation defining this process is represented by

\begin{equation} 
\mathcal{L}_{NCS} = \text{NCS}(f^{s}_{cls}, \hat{f}^{a}_{cls}) + \text{NCS}(f^{s}_{cls}, \hat{f}^{b}_{cls})
\end{equation}
where $\text{NCS}$ stands for the Negative Cosine Similarity loss with a stop-gradient operation (\texttt{stopgrad}) \cite{chen2021exploring}, given as

\begin{equation} 
    \text{NCS}(p, \texttt{stopgrad}(z)) = - \lnorm{p}{\cdot}\lnorm{\texttt{stopgrad}(z)}.
\end{equation}

The stop-gradient operation is applied to $\hat{f}^{a}_{cls}$ and $\hat{f}^{b}_{cls}$ to prevent the reference encoders and the MLP projections from being updated with respect to the NCS loss. Therefore, the projected $x_{\text{class}}$ vectors are considered as ground-truth representations for the superimposed encoder and are only updated using the CE loss. Notably, we propose three variants of our approach. The first variant, TransformAR, represents the framework (Figure \ref{fig:arch}), but it does not include the MLPs designed to enhance learning representation capability. The second variant, TransformAR-KD, incorporates knowledge distillation to improve learning ability. The final variant, TransformAR-KD+, extends the previous one by also feeding $x_{\text{class}}$ of each encoder to the quality-aware decoder.

\subsection{Quality-Aware Decoders}
\label{ssec:quality_decoders}
We employ the transformer decoder \cite{vaswani2017attention} used for machine translation. However, we remove the masked self-attention mechanism, since our goal is not to predict the next token. Although in our task, we retain the core idea, which is the use of the Cross-Attention (CA) mechanism. In our setup, queries originate from the reference representations, while keys and values are derived from the shift representations. This alignment process facilitates the generation of a sequence of quality embedding vectors by aligning pertinent information from the reference representations with relevant information from the shift representations caused by distortions.

As depicted in Figure \ref{fig:arch}, $\mathcal{G}^{as}(\cdot)$ and $\mathcal{G}^{bs}(\cdot)$ represent the quality decoder of the superimposed image $I^s$ based on the reference foreground image $I^a$ and based on the reference background $I^b$, respectively. Each transformer decoder is similar to the transformer encoder with the exception of the use of the CA mechanism instead of self-attention.  CA is calculated by
\begin{align}
    \text{CA}(q_{f^{i}}, k_{d^{is}}, v_{d^{is}}) = \textit{softmax}\left(\frac{q_{f^{i}}k^{T}_{d^{is}}}{\sqrt{d_k}}\right) v_{d^{is}}, 
\end{align}
where $q_{f^{i}}$ denotes the queries from $f^i$, which are our reference embedding vectors. Keys $k_{d^{is}}$ and values $v_{d^{is}}$ are calculated from the shift representations $d^{is}$, and $d_k$ represents the dimension of the key vectors. Following this, a normalization layer is applied to the output embedding vectors with a skip connection to preserve the distance information caused by distortions. The output is then projected using a feed-forward neural network, producing a sequence of embedding vectors $g^{is} = \mathcal{G}^{is}(f^i, d^{is})$ containing quality features, where $i \in \{a, b\}$ (Figure \ref{fig:enc-dec}). 

By adopting this approach to discern quality features between reference image embedding vectors and distortions, or shift embedding vectors, attention scores serve as crucial indicators. High attention scores between corresponding elements signify minimal distortion, and suggesting little disruption to the content. Conversely, low attention scores denote significant distortion at specific positions, indicating pronounced content alteration. In addition, Content-aware encoders represent both images as plausibly similar if the applied distortion is minimal. This helps the model represent the quality information based on the shift caused by the distortions $D(.)$ applied on the AR image as well as the visual confusion caused by overlaying the AR image onto the background at a certain threshold.

\subsection{Regression Modules}
\label{ssec:reg}
Finally, the model incorporates two regression modules, denoted as $\mathcal{R}^{as}(\cdot)$ and $\mathcal{R}^{bs}(\cdot)$, which are applied to $g^{as}$ and $g^{bs}$ respectively, to generate quality scores for each patch $x_i$ from the superimposed image $I^s$. The first set of scores is computed as $h^{as} = \mathcal{R}^{as}(g^{as})$, while the second set is computed as $h^{bs} = \mathcal{R}^{bs}(g^{bs})$. These scores are then combined using a linear layer with learnable parameters $W^{as}$ and $W^{bs}$ (without a bias term), resulting in two final scores $S^{as} = W^{as}h^{as}$ and $S^{bs} = W^{bs}h^{bs}$ for the superimposed image $I^s$. 

The next step involves aggregating these scores using a hyperparameter $\zeta$, which yields the predicted Mean Opinion Score (pMOS) according to the following formula

\begin{equation}
    \textit{pMOS} = \zeta S^{as} + (1 - \zeta)S^{bs}.
\end{equation}
Here, $\zeta$ is empirically set to 0.51 after iterative experimentation and performance evaluation.


\section{Training Implementation}
\label{sec:train_impl}
The training process involves several crucial elements, such as selecting the appropriate loss function and mitigating overfitting by employing regularization techniques, specifically elastic net regularization \cite{zou2005regularization} and label smoothing for regression problem.

\subsection{Objective Function}
\label{subsec:huber}
In addition to the loss function used for classification and enhancing the learned representation for the superimposed encoder, as described in Section \ref{ssec:enhc_rep}. We have opted for the Huber loss \cite{huber1992robust} as an objective function to assess the predicted quality score. This loss function is commonly employed in machine learning and regression problems. It combines the properties of the MSE and the Mean Absolute Error (MAE), offering a balanced approach between robustness to outliers and sensitivity to small errors. The Huber loss equation is expressed as 

\begin{align}
\mathcal{L}_{H}(q, \hat{q}, \delta) = 
\begin{cases} 
\frac{1}{2}(q - \hat{q})^2 & \text{for } |q - \hat{q}| \leq \delta \\
\delta(|q - \hat{q}| - \frac{1}{2}\delta) & \text{otherwise}
\end{cases}
\end{align}
where, $q$ represents the target MOS, and $\hat{q}$ represents  pMOS. $\delta$ serves as a hyperparameter determining the threshold at which the loss transitions from quadratic behavior for small residuals to linear behavior for larger residuals \cite{huber1992robust}, set to one empirically.

\subsection{Regularization}
To address the overfitting issue, we employ a regularization technique known as elastic net \cite{zou2005regularization}. This method combines Lasso and Ridge regularization approaches, effectively preventing overfitting and serving as a feature selection mechanism in regression models. The regularization loss function is defined as

\begin{equation}
 \mathcal{L}_{R} = \alpha\|\theta_{\mathcal{R}}\|_1 + (1 - \alpha)\|\theta_{\mathcal{R}}\|^2_2
\end{equation}
where $\theta_{\mathcal{R}}$ represents the learnable parameters of the regressor modules, and $\alpha$ is a hyperparameter set to $0.7$ for controlling the strength of $l_1$ and $l_2$, respectively. By combining losses, we formulate the comprehensive loss function

\begin{equation}
 \mathcal{L} = \lambda_0\mathcal{L}_{H} + \lambda_1\mathcal{L}_{NCS} + \lambda_2\mathcal{L}_{CE} + \lambda_3\mathcal{L}_{R}.
\end{equation}
In this equation, $\lambda_i$ for $i={0, 1, 2}$ are set to one except for $\lambda_3$, which is set to $0.05$. These hyperparameters  regulate the influence of each loss term on the overall objective function.

\subsection{Label Smoothing}
\label{ssec:label_smooth}
In our final step, we employ label smoothing, a regularization technique commonly used in classification tasks. Traditionally, label smoothing has not been applied to regression problems, as regression involves predicting continuous values rather than probability distributions. However, in scenarios where labels contain inherent noise, particularly in cases with limited data, such noise can significantly impact the performance of the model. Overfitting may occur as the model learns patterns specific to the noise present in the training set. To mitigate this, when the model begins to overfit, we introduce some level of noise to the labels, such as

\begin{align} \label{eq:noise}
    MOS_{\epsilon} = MOS + \eta_n \epsilon
\end{align}
In Equation \eqref{eq:noise}, $\eta_n$ is a random value drawn from a uniform distribution ranging from -1 to 1. This indicates if the introduced noise $\epsilon$ is added or subtracted from the actual scores. $\epsilon$ is a random number sampled from a normal distribution with a mean of 0 and variance of 1.

\begin{figure}[t]
  \centering
    \setlength{\belowcaptionskip}{-10pt}
    \begin{tabular}{c}
      \includegraphics[width=0.90\linewidth]{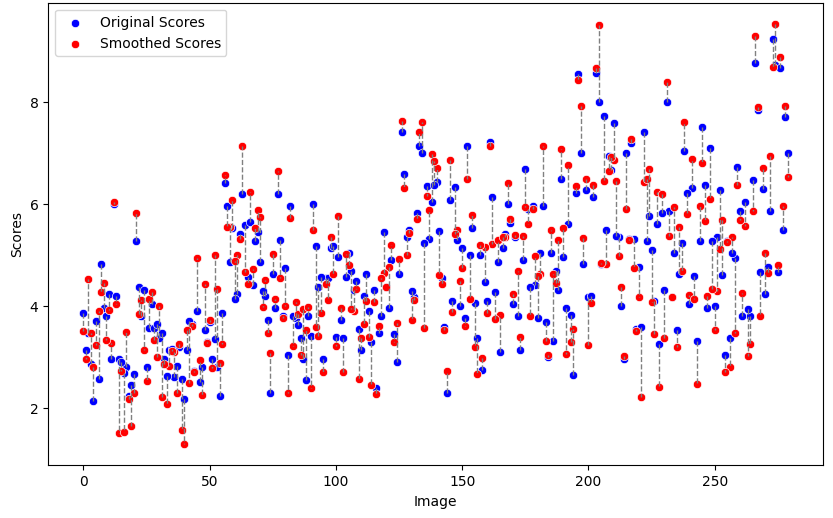}\\
      \small (a)\\
      \includegraphics[width=0.90\linewidth]{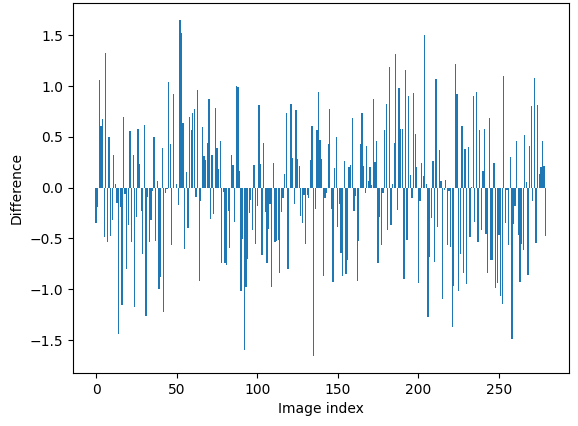}\\
      \small (b)
    \end{tabular}
  \caption{Impact of label smoothing: (a) Scores before (blue dots) and after (red dots), and (b) Difference between scores.}
  \label{fig:diff_score_smooth}
\end{figure}

Figure \ref{fig:diff_score_smooth} illustrates the effects of employing the label smoothing technique, starting with Figure \ref{fig:diff_score_smooth}-(a) which shows the MOS before (blue dots) and after (red dots) applying this technique. Figure \ref{fig:diff_score_smooth}-(b) depicts the differences between the original scores and the smoothed ones, indicating that some scores were altered significantly while others experienced minimal changes.

\subsection{Configuration details}
To train the model, we employed the AdamW optimizer \cite{loshchilov2017decoupled} with a learning rate set to $1e^{-4}$ for the quality-aware decoders, the regressors, and the MLPs. For fine-tuning the content-aware encoders and to retain prior knowledge and adapt seamlessly to our dataset, we used a small learning rate $1e^{-5}$. Training was conducted with a batch size of 32 images for 250 epochs. Furthermore, we incorporated a learning rate scheduler, specifically ``reduce on plateau'', which decreased the learning rate if there was no improvement in model performance over 20 consecutive epochs. This adjustment aimed to encourage a more meticulous exploration of the parameter space. The implementation was carried out using PyTorch on an NVIDIA Tesla V100S-PCIE-32GB GPU.

\subsection{ARIQA Dataset}
\label{ssec:dataset}

To assess the effectiveness of our proposed approach, we conducted experiments on the ARIQA dataset, which consists of 560 superimposed images along with their associated MOS. To ensure a fair comparison, we used the same five folds employed by Duan et al. \cite{duan2022confusing}. Additionally, to better evaluate our approach, we divided the dataset into another 50 folds following the same strategy, denoted as $\mathcal{X}$. Each fold $i$ was randomly partitioned into 280 training samples, denoted as $\mathcal{T}_i$ and 280 testing samples, denoted as $\mathcal{S}_i$, ensuring that scenes were not repeated between the training and testing sets, as detailed in Equations \eqref{eq:ds} and \eqref{eq:ds2} where $\mathcal{D}$ refers to the entire ARIQA dataset.
\begin{align} 
    \label{eq:ds}
    \mathcal{D} &= \{[I^a_{i}, I^b_{i}, I^s_{i}, MOS_i]\}_{i=1}^{560} \\ 
    \label{eq:ds2}
    \mathcal{X} &= \{(\mathcal{T}_i, \mathcal{S}_i) \ | \ \mathcal{T}_i \cap \mathcal{S}_i = \emptyset\}_{j=1}^{50} \text{where} \ \bigcup_{j=1}^{50} \mathcal{S}_j = \mathcal{D}
\end{align}

\begin{figure}[!ht]
    \centering
    \begin{minipage}[b]{0.33\linewidth}
      \centering
      \includegraphics[width=\linewidth]{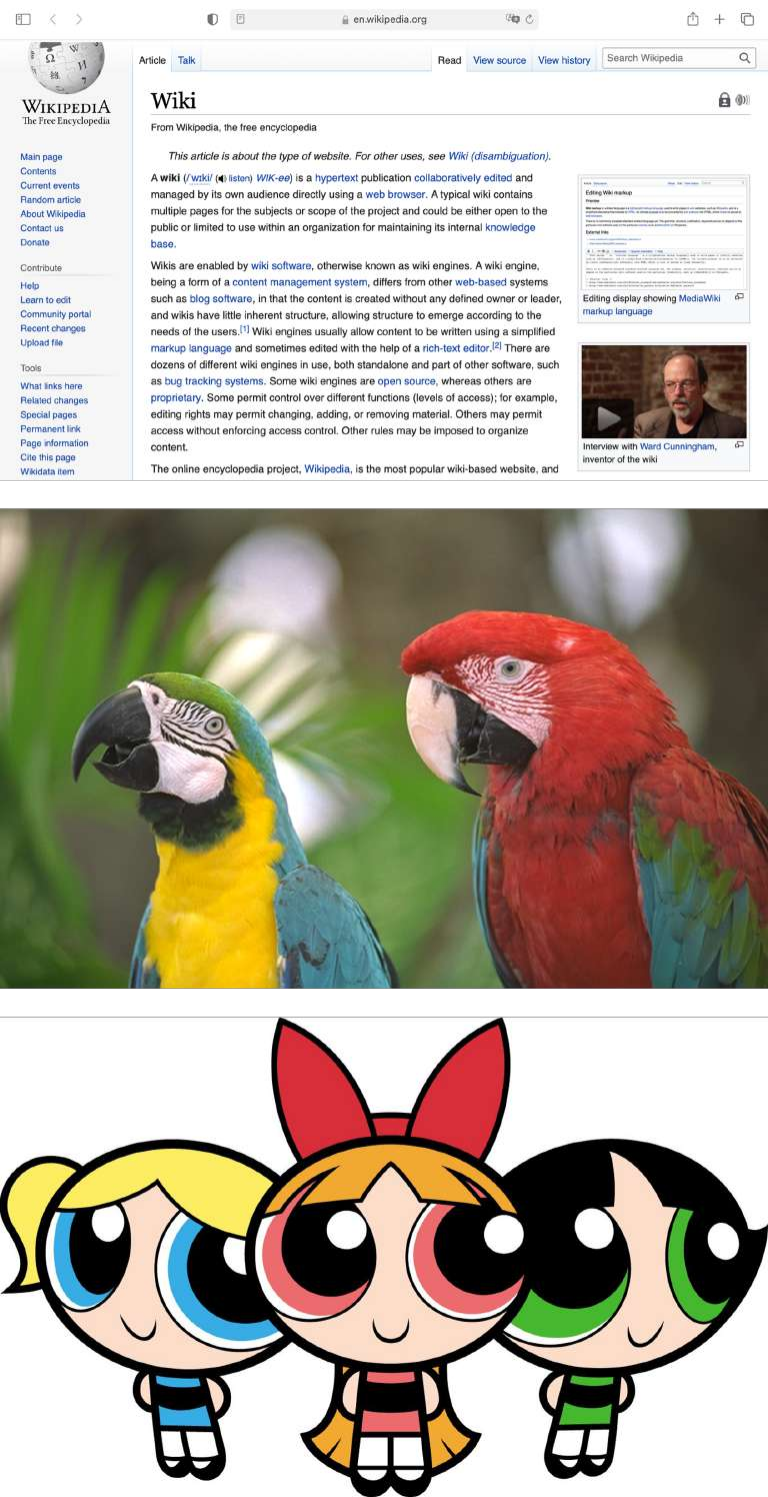}
      \caption*{FG images $I^a$}
    \end{minipage}%
    \hfill
    \begin{minipage}[b]{0.33\linewidth}
      \centering
      \includegraphics[width=\linewidth]{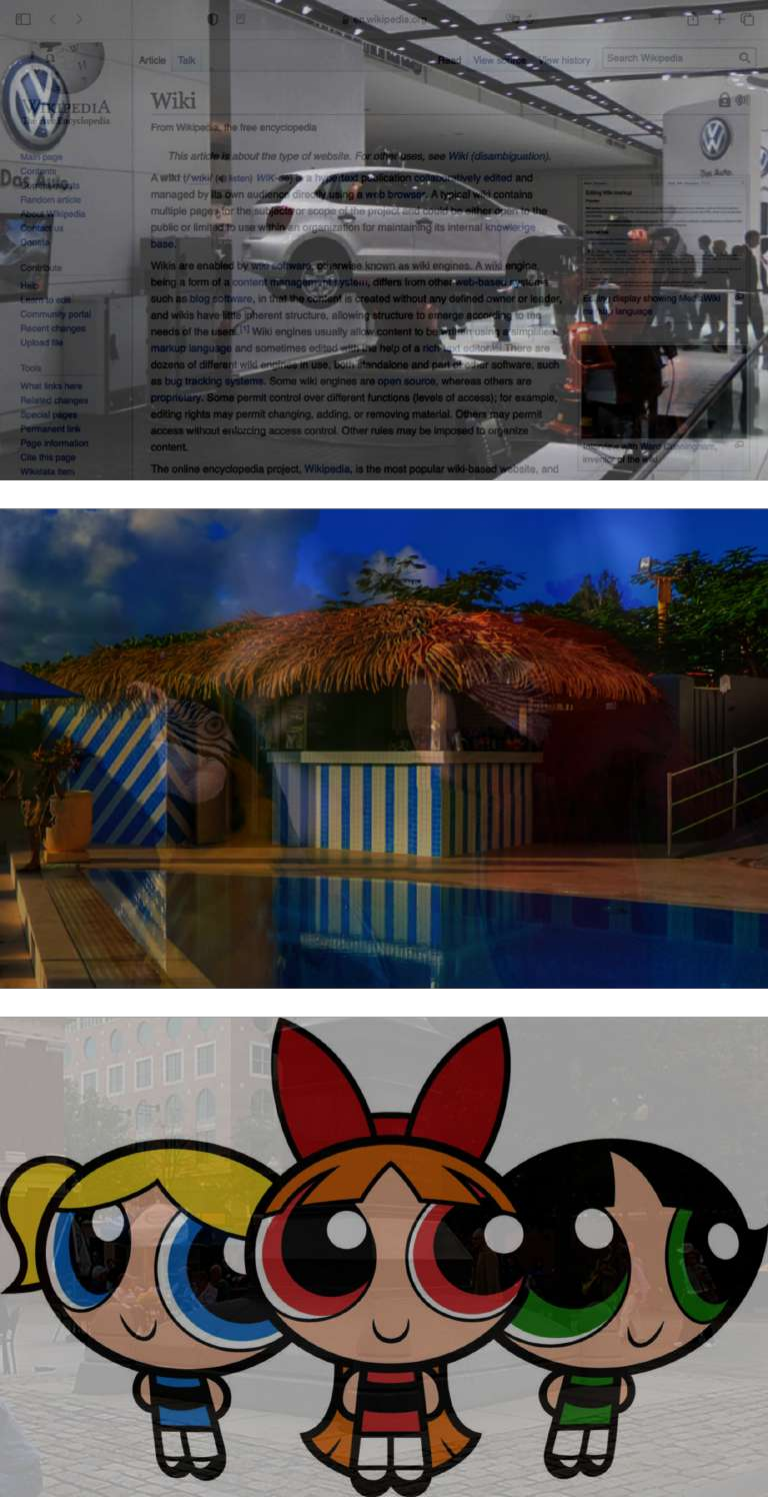}
      \caption*{Distorted images $I^s$}
    \end{minipage}%
    \hfill
    \begin{minipage}[b]{0.33\linewidth}
      \centering
      \includegraphics[width=\linewidth]{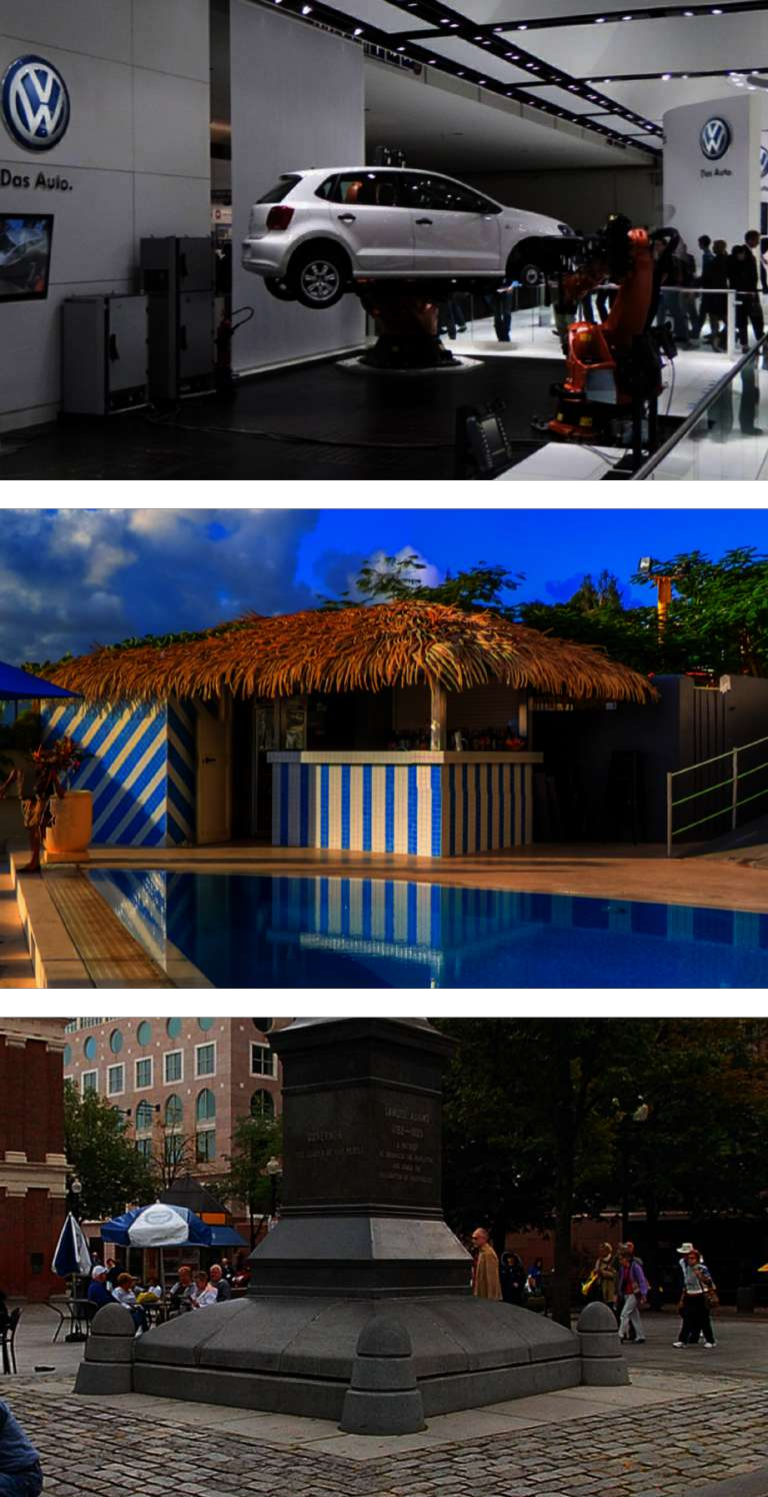}
      \caption*{BG images $I^b$}
    \end{minipage}
    \caption{Illustration of reference and superimposed (distorted) images from the used dataset \cite{duan2022confusing}.}
    \label{fig:ariqa_samples}
\end{figure}

The ARIQA dataset consists of 20 omnidirectional background images, evenly split between indoor and outdoor scenes, along with 20 reference AR images falling into three categories (web, natural images, and graphics), all with a resolution of $1440 \times 900$. Figure \ref{fig:ariqa_samples} shows a sample from each category, in which each AR image was subjected to six degraded versions using JPEG compression, image scaling, contrast adjustment, and the inclusion of visual confusion as a distortion factor \cite{duan2022confusing}. A total of 560 stimuli are chosen from all possible combinations for a subjective experiment involving 23 participants using HTC VIVE Pro Eye VR headsets. As reported by Duan et al. \cite{duan2022confusing}, subjects rate the perceived quality on a 10-point scale using a single-stimulus strategy across 20 scenarios, each with seven superimposed levels, presented randomly. The experiment aimed to assess how different degradation processes and visual confusion affect quality in AR scenarios. The viewport containing the foreground image superimposed on the background is saved along with its associated quality score. Figure \ref{fig:viewport} illustrates one of the many AR scenarios used during the subjective test.

\section{Results and discussions}
\label{sec:res_disc}
This section presents a comprehensive analysis of our proposed approach, wherein we calculate various metrics across all model variants. A five-parameter logistic function 
\begin{equation}\label{eq:logistic}
  Q'=\beta_{1}(\frac{1}{2}-\frac{1}{1+e^{\beta_{2}(Q-\beta_{3})}})+\beta_{4}Q+\beta_{5},
\end{equation}
was employed to model the relationship between the objective quality scores and the best-fitting quality scores. In Equation \eqref{eq:logistic} $Q$ and $Q'$ represent the objective and fitted quality respectively, and $\beta_{i}$s ($i=1,2,3,4,5$) are the parameters adjusted during the evaluation process.
Four evaluation metrics are used to measure the consistency between the subjective ratings provided as ground truth and the quality scores obtained through fitting, namely, Spearman Rank-order Correlation Coefficient (SRCC), Kendall Rank-order Correlation Coefficient (KRCC) and Pearson Linear Correlation Coefficient (PLCC). We conduct comparisons with state-of-the-art methods, and  delve into an ablation study to discern the effectiveness of individual components inherent in our objective metric.

\begin{table}[h]
  \centering
    \setlength{\tabcolsep}{3pt}
    \fontsize{9pt}{\baselineskip}\selectfont
  \caption{Comparison of our results with state-of-the-art performance. (The bold numbers represent the best results.)}
  \label{tab:sota}
  \begin{tabular}{lccc}
    \toprule
    Model $\backslash$ Criteria & SRCC$\uparrow$ & KRCC$\uparrow$ & PLCC$\uparrow$ \\ 
    \midrule
    PSNR & 0.2900 & 0.1977 & 0.3091 \\
    SSIM \cite{wang2004image} & 0.2521 & 0.1698 & 0.2850 \\
    MS-SSIM \cite{wang2003multiscale} & 0.1019 & 0.0687 & 0.1491 \\
    GMSD \cite{xue2013gradient} & 0.2680 & 0.1842 & 0.3261 \\
    MDSI \cite{nafchi2016mean} & 0.2931 & 0.1978 & 0.3023 \\
    IW-SSIM \cite{wang2010information} & 0.3499 & 0.2410 & 0.3635 \\
    DSS \cite{balanov2015image} & 0.3587 & 0.2448 & 0.3906 \\
    HaarPSI \cite{reisenhofer2018haar} & 0.3883 & 0.2677 & 0.4243 \\
    VIF \cite{sheikh2006image} & 0.4848 & 0.3417 & 0.5175 \\
    VSI \cite{zhang2014vsi} & 0.5632 & 0.4021 & 0.5802 \\
    SR-SIM \cite{zhang2012sr} & 0.6093 & 0.4282 & 0.6212 \\
    FSIM \cite{zhang2011fsim} & 0.6124 & 0.4419 & 0.6320 \\
    \midrule 
    LPIPS \cite{zhang2018unreasonable} & 0.7624 & 0.5756 & 0.7591 \\ 
    CFIQA \cite{duan2022confusing} & 0.7787 & 0.5863 & 0.7695 \\ 
    ARIQA \cite{duan2022confusing} & 0.7902 & 0.5967 & 0.7824 \\ 
    ARIQA+ \cite{duan2022confusing} & 0.8124 & 0.6184 & 0.8136 \\ 
    TransformAR \cite{asekhri} & 0.7765 & 0.5868 & 0.7749 \\ 
    TransformAR-KD  & 0.8390 & 0.6503 & 0.8383 \\ 
    TransformAR-KD+  & \textbf{0.8411} & \textbf{0.6538} & \textbf{0.8416} \\ 
    \midrule
    TransformAR \cite{asekhri} (50-folds) & 0.8267 & 0.6359 & 0.8251 \\ 
    TransformAR-KD (50-folds) & 0.8563 & 0.6698 & \textbf{0.8581} \\ 
    TransformAR-KD+ (50-folds) & \textbf{0.8566} & \textbf{0.6712} & 0.8580 \\ 
    \bottomrule
  \end{tabular}
\end{table}

\subsection{Comparison with State-of-the-Art}

Table \ref{tab:sota} presents the state-of-the-art comparison against classical methods as well as the results reported in the work of Duan et al. \cite{duan2022confusing}. For the classical FR-IQA methods, we compute the overall quality by employing Support Vector Regression (SVR) \cite{chang2011libsvm} \cite{kastryulin2022piq} to measure the similarity between the distorted image $I^s$ and the reference AR image $I^a$, and the similarity between the distorted image $I^s$ and the background image $I^b$. These can be expressed as:
\begin{equation}
  Q_{overall} = SVR_{rbf}(FR(I^s, I^a), FR(I^s, I^b)),
\end{equation}
where $Q_{overall}$ is the overall predicted quality, and $SVR_{rbf}$ stands for support vector regression using the radial basis function kernel. We compute the average of these metrics across the five folds, as well as across  50 folds. Our approaches, TransformAR-KD and TransformAR-KD+, outperform all previous methods across all performance metrics on the five folds used by \cite{duan2022confusing}. Specifically, TransformAR-KD achieves improvements of 3.53\%, 5.73\%, and 3.44\% over the state-of-the-art model ARIQA+ for SRCC, PLCC, and KRCC, respectively. TransformAR-KD+ performs even better, highlighting the significance of the $x_{\text{class}}$ token in capturing global information for high-semantic tasks. While TransformAR performs better than LPIPS, it is outperformed by ARIQA and ARIQA+, demonstrating the limitations of not fine-tuning the encoders to capture detailed content information about the input images. Classical methods, designed for natural images, struggle to predict quality in a way that aligns with the HVS, emphasizing the need for more advanced approaches to effectively handle visual confusion.
In the 50 folds, which provide a more precise assessment, TransformAR shows good performance, surpassing all previous methods. Fine-tuning our reference content-aware encoders and minimizing the NCS loss to align the superimposed representation with the reference ones have resulted in better representations by learning both the category of the AR image and the nature of the background leading to significantly better performance across all metrics.

This highlights the effectiveness of our proposed approach for capturing precise semantic information, which includes features about the nature of the input images. This enables the generation of more representative shift embedding vectors, which are then used for generating quality representations. In contrast, the ARIQA and LPIPS models, which are mainly based on CNNs, are adept at learning hierarchical features but suffer from a limited receptive field. This limitation is not present in ViTs, showcasing the superiority of these models in capturing long-range dependencies even in the presence of visual confusion.

\begin{table*}[ht!]
    \centering
    \setlength{\tabcolsep}{3pt}
    \fontsize{7.2pt}{\baselineskip}\selectfont
    \caption{Ablation study results by training the model on five folds from $\mathcal{X}$. The bold numbers indicate the best results, while the red numbers represent the worst results.}
    \label{tab:ablation_all}


    \begin{tabular}{l cccc c cccc c cccc}
        \toprule
        Model & \multicolumn{4}{c}{TransformAR \cite{asekhri}} & & \multicolumn{4}{c}{TransformAR-KD} & & \multicolumn{4}{c}{TransformAR-KD+} \\
        \cline{2-5} \cline{7-10} \cline{12-15}
        \rule{0pt}{2.6ex}
        Setting $\backslash$ Criteria & SRCC$\uparrow$ & KRCC$\uparrow$ & PLCC$\uparrow$ & RMSE$\downarrow$ & & SRCC$\uparrow$ & KRCC$\uparrow$ & PLCC$\uparrow$ & RMSE$\downarrow$ & & SRCC$\uparrow$ & KRCC$\uparrow$ & PLCC$\uparrow$ & RMSE$\downarrow$\\
        \midrule
        1 - w/o decoder & 0.6161 & 0.4408 & 0.6242 & 1.2561 & & 0.5688 & 0.4052 & 0.5832 & \textcolor{red}{2.6103} & & 0.4361 & 0.3014 & \textcolor{red}{0.4236} & \textcolor{red}{2.6449} \\
        2 - w/o $l_1$-distance & \textcolor{red}{0.4365} & \textcolor{red}{0.2960} & \textcolor{red}{0.4443} & \textcolor{red}{1.5248} & & \textcolor{red}{0.4081} & \textcolor{red}{0.2833} & \textcolor{red}{0.4629} & 1.3928 & & \textcolor{red}{0.4124} & \textcolor{red}{0.2824} & 0.4542 & 1.5008 \\
        3 - w/o label smoothing & 0.8269 & 0.6361 & 0.8221 & 1.0552 & & 0.8368 & 0.6476 & 0.8375 & 0.9678 & & 0.8408 & 0.6540 & 0.8453 & 0.9651 \\
        4 - w/o elastic net & 0.8427 & 0.6567 & 0.8471 & \textbf{0.9014} & & 0.8547 & 0.6686 & 0.8618 & 0.9212 & & 0.8586 & 0.6758 & \textbf{0.8655} & 0.8592 \\
        5 - w/o Huber loss & 0.8374 & 0.6525 & 0.8443 & 0.9444 & & 0.8506 & 0.6617 & 0.8507 & 0.8507 & & 0.8541 & 0.6712 & 0.8610 & 0.9799 \\
        all combined & \textbf{0.8461} & \textbf{0.6582} & \textbf{0.8481} & 0.9168 & & \textbf{0.8637} & \textbf{0.6791} & \textbf{0.8685} & \textbf{0.9034} & & \textbf{0.8593} & \textbf{0.6762} & 0.8602 & \textbf{0.8610} \\
        \bottomrule
    \end{tabular}
\end{table*}

\subsection{Ablation Study}
In this section, we empirically study the effect of the major components on our proposed method using only five folds from $\mathcal{X}$. In each experiment, we systematically removed one component to analyze its impact on the overall results. The chosen components for ablation include the quality-aware decoder, $l_1$-distance for calculating the shift representations, label smoothing, elastic net regularization for tackling the overfitting problem, and the use of MSE instead of the Huber loss. The final experiment involves using all components combined, which represents the proposed approaches. Note that all experiments are conducted for all three model's variants.

\subsubsection{The effect of the quality-aware decoder}
In the first experiment, we omitted the use of the quality-aware decoder. The results, as depicted in Table \ref{tab:ablation_all}, exhibited a noticeable decline in all model variants. This highlights the importance of this module, which is responsible for generating the set of quality vectors $g_{as}$ and $g_{bs}$ by facilitating communication between shift and reference vectors through the cross-attention mechanism.

\subsubsection{The effect of the shift representations}
In the second experiment, we excluded the calculation of the shift representations $d_{as}$ and $d_{bs}$ using $l_1$-distance. Instead, we directly fed the set of distorted vectors $f^s$ to the decoder alongside the set of reference vectors $f^i$, where $i=\{a, b\}$. It became evident that the model failed to capture quality features due to the absence of distortion information. This resulted in a significant performance decrease of nearly 50\% across all metrics. A comparative analysis of the model's variants revealed that TransformAR-KD and TransformAR-KD+ performed worse than TransformAR, where enhancement of learned representations as we discussed in Section \ref{ssec:enhc_rep} was not involved. This can be attributed to the fact that TransformAR's content-aware encoders are frozen, serving as deterministic functions. Thus, even images with slight difference get embedded in distinct ways while similar images get represented in the same manner, potentially aiding the decoder more effectively than fine-tuning the content-aware encoders as in TransformAR-KD and TransformAR-KD+.

\subsubsection{The effect of overfitting mitigation}
\begin{figure}[!t]
    \centering
    \includegraphics[width=\columnwidth]{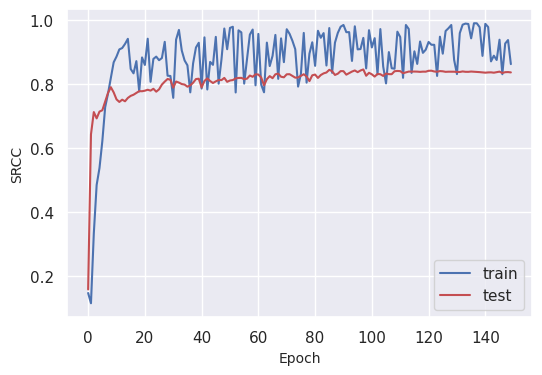}
    \caption{Evolution of SRCC performance across epochs within a single fold.}
    \label{fig:srcc2}
\end{figure}

For the third and fourth experiments, we aimed to evaluate the effects of employing label smoothing and elastic net regularization to address the overfitting problem. Beginning with the exclusion of label smoothing, a significant decrease was observed in terms of SRCC, KRCC, and PLCC, achieving $0.8269$, $0.6361$, and $0.8221$, respectively. This decline can be attributed to the model's tendency to become overly confident in predicted scores during training, potentially learning patterns influenced by intrinsic noise present in the actual MOS data. The latter leads to the overfitting problem. This phenomenon was particularly noticeable in the initial epochs (Figure \ref{fig:srcc2}). However, introducing artificial noise, as explained in Section \ref{ssec:label_smooth}, and prolonging training duration helped improve results (Table \ref{tab:ablation_all}). In the fourth experiment, despite the role of elastic net regularization in preventing overfitting, removing it had a minimal effect on the results. However, this minimal effect is still important, especially when dealing with small datasets. Penalizing the regression weights helps the model select the prominent features from the quality embedding vectors.

\subsubsection{The effect of the Huber loss}
The final experiment in the ablation involved replacing the Huber loss with the MSE loss, leading to a notable drop in all metrics. This decline was particularly prominent in the case of SRCC across all model variants, with a slight decrease observed in the rest of the metrics as well. This emphasizes the importance of the Huber loss due to its robustness against outliers as described in Section \ref{subsec:huber}.
\begin{table*}[ht!]
    \centering
    \renewcommand{\arraystretch}{1.25}
    \setlength{\tabcolsep}{3pt}
    \fontsize{7.2pt}{\baselineskip}\selectfont
    \caption{Performance comparison across the five folds for each mixing value $\sigma$. The bold numbers represent the best performance for each metric across the different models under the same mixing value.}
    \label{tab:ablation_mixing_all}

    \begin{tabular}{l cccc c cccc c cccc}
        \toprule
        Model & \multicolumn{4}{c}{TransformAR \cite{asekhri}} & & \multicolumn{4}{c}{TransformAR-KD} & & \multicolumn{4}{c}{TransformAR-KD+} \\
        \cline{2-5} \cline{7-10} \cline{12-15}
        \rule{0pt}{2.6ex}
        Mixing value $\backslash$ Criteria & SRCC$\uparrow$ & KRCC$\uparrow$ & PLCC$\uparrow$ & RMSE$\downarrow$ & & SRCC$\uparrow$ & KRCC$\uparrow$ & PLCC$\uparrow$ & RMSE$\downarrow$ & & SRCC$\uparrow$ & KRCC$\uparrow$ & PLCC$\uparrow$ & RMSE$\downarrow$ \\
        \midrule
        $\sigma_1 = 0.26$ & 0.7682 & 0.6177 & 0.7922 & 0.8162 & & 0.8625 & 0.7018 & 0.8674 & 0.6801 & & \textbf{0.9042} & \textbf{0.7432} & \textbf{0.9126} & \textbf{0.4654} \\
        $\sigma_2 = 0.42$ & 0.8649 & 0.7074 & 0.8647 & \textbf{0.5726} & & 0.8717 & 0.7085 & 0.8700 & 0.6247 & & \textbf{0.8736} & \textbf{0.7162} & \textbf{0.8782} & 0.6048 \\
        $\sigma_3 = 0.58$ & 0.8895 & 0.7301 & \textbf{0.8909} & \textbf{0.5054} & & \textbf{0.8957} & \textbf{0.7437} & 0.8899 & 0.5312 & & 0.8660 & 0.7005 & 0.8761 & 0.6761 \\
        $\sigma_4 = 0.74$ & 0.8069 & 0.6459 & 0.8308 & 0.6955 & & \textbf{0.9045} & \textbf{0.7501} & \textbf{0.9055} & \textbf{0.4731} & & 0.8998 & 0.7463 & 0.8930 & 0.5234 \\
        \bottomrule
    \end{tabular}
\end{table*}

\subsubsection{The effect of the mixing thresholds}
Table \ref{tab:ablation_mixing_all} presents a comparative analysis of performance across three proposed model variants over five folds, for each mixing value ($\sigma$) used in the dataset \cite{duan2022confusing}, where $\sigma \in [0.26, 0.42, 0.58, 0.74]$. This investigation aims to elucidate the influence of the mixing value on overall performance. Obviously, TransformAR-KD and TransformAR-KD+ outperform the base model. This superiority can be attributed to the inclusion of reference image training, which enables the encoders to treat input reference images differently. Notably, TransformAR achieves superior results in terms of PLCC and RMSE when $\sigma = 0.58$. However, a closer examination between TransformAR-KD and TransformAR-KD+ reveals nuanced performance variants; TransformAR-KD+ excels when $\sigma < 0.5$, while TransformAR-KD performs better when $\sigma > 0.5$. This distinction can be clarified by considering the impact of $\sigma$ on the transparency of the AR image $I^a$, as described in Equation \eqref{eq:mixing}. A smaller $\sigma$ implies greater transparency, therefore allowing the quality-aware decoders to preserve content features from the AR image by leveraging the $x_{\text{class}}$ feature representation $f^{a}_{cls}$, leading to better performance. Conversely, for larger $\sigma$ values, the AR image becomes more prominent, indicating that the distilled knowledge encompasses ample information about the reference image.

\section{Qualitative analysis}
\label{sec:qual_analysis}
In this section, we focus on the qualitative aspects of our study. We present attention maps generated by the model's encoders. Then, we use UMAP \cite{mcinnes2018umap} to visually showcase the model's capability in separating classes based on the learned features at the $x_{\text{class}}$ vector.

\begin{figure*}[!ht]
    \centering
    \includegraphics[width=\textwidth]{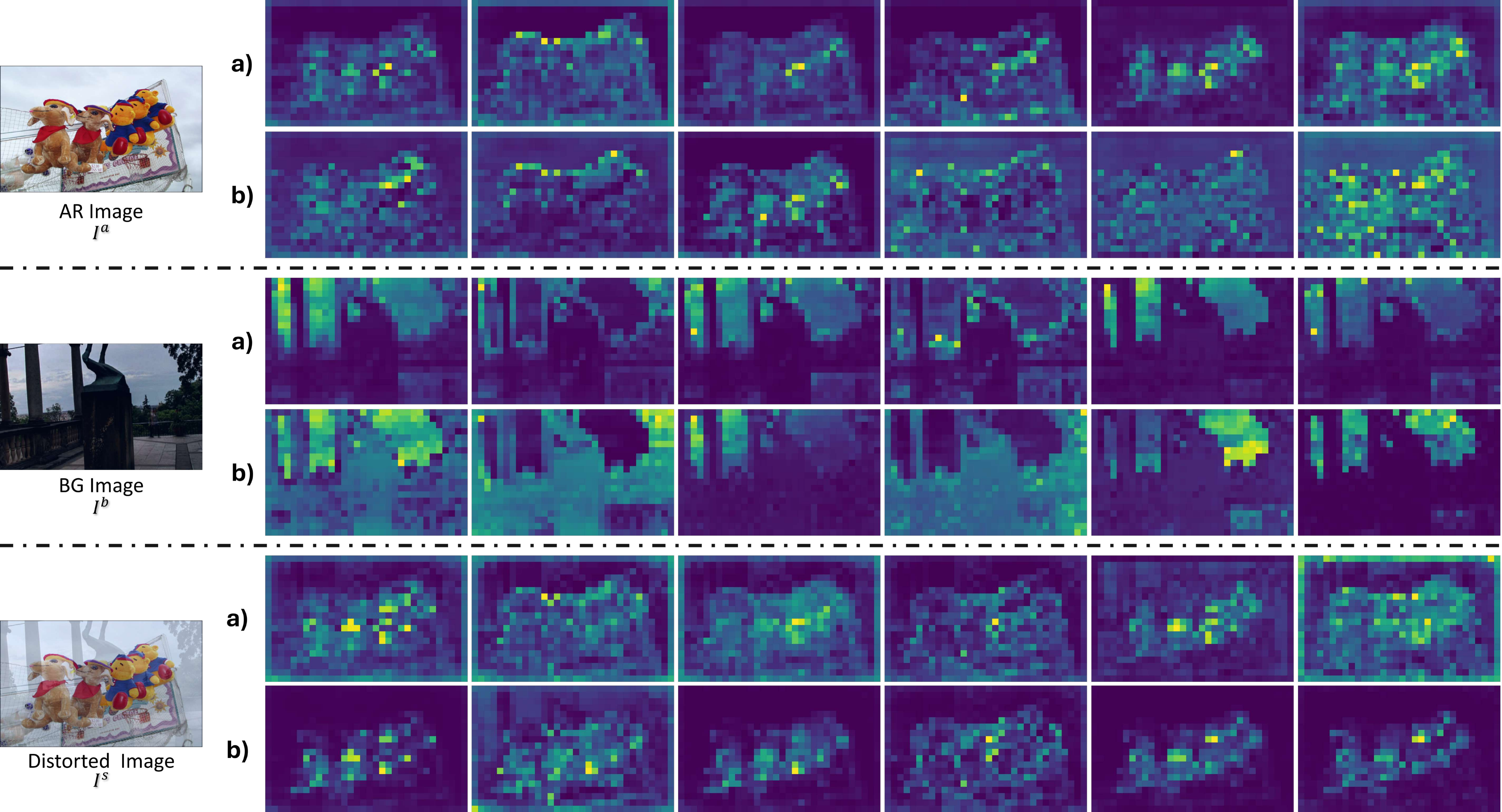}
    \caption{Self-attention from the encoders of both, \textbf{(a)} TransformAR \cite{asekhri} and \textbf{(b)} TransformAR-KD with $16 \times 16$ patches. We look at the self-attention of the $x_{\text{class}}$ on the heads of the last encoder block. For the TransformAR-KD as the encoders are trained, therefore, this token is  attached to classifiers (AR image and background image) heads to predict to which category the images belong. The colors correspond to the intensity of self-attention, with darker colors representing low attention and brighter colors representing high attention.}
    \label{fig:att_transformar}
\end{figure*}

\subsection{Attention maps visualization}
Attention maps of the $x_{\text{class}}$ token provide insights into the regions the model predominantly focuses on, guided by the objective functions. We investigate the self-attention mechanisms of the content-aware encoders in TransformAR and TransformAR-KD models to understand the impact of aligning distorted and reference representations by maximizing cosine similarity (Figure \ref{fig:arch}). Analyzing the AR image $I_a$, the $x_{\text{class}}$ token in TransformAR’s encoders Figure \ref{fig:att_transformar}-\textbf{(a)} shows focused attention on key regions of $I^a$. These attention maps reveal information pertinent to the semantic segmentation of the image \cite{caron2021emerging}. With the integration of knowledge distillation, Figure \ref{fig:att_transformar}-\textbf{(b)} illustrates that TransformAR-KD’s encoders capture more diverse semantic regions of $I^a$, showing that different heads emphasize varied semantic details, reflecting the enhanced representation learning after fine-tuning on the ARIQA dataset. This results in distinct attention patterns in visually complex regions compared to the baseline TransformAR.

For the background image $I^b$, the TransformAR's $x_{\text{class}}$ predominantly reaches the regions where the sky is located and that is across all heads, disregarding other objects. This behavior is attributed to the complexity of outdoor scenarios for the frozen encoder. This issue was not the case for TransformAR-KD, where it becomes evident that by supplying the model with information about the background's nature, whether indoor or outdoor, the $x_{\text{class}}$ token focuses on major components without being overly influenced by the sky, which typically appears brighter compared to other objects.

Finally, when it comes to the superimposed image $ I^s $ that has been distorted by classical distortions and also the visual confusion, TransformAR's $x_{\text{class}}$ token exhibits attention not only towards the AR image but also towards the background in an incomprehensible manner, especially at the image borders where the model was paying attention to these areas as well. This underscores the influence of visual confusion, whereby these confused images deviate from the actual data upon which the model was trained (ImageNet), thereby causing data drift issues. This problem does not appear in the case of the TransformAR-KD, where the model is adapted to the ARIQA dataset. The model predominantly focuses on the AR object, and the attention maps are appropriately distributed across various parts of the AR image.

\begin{figure*}[!ht]
    \centering
    \begin{minipage}{\textwidth}
        \centering
        \includegraphics[width=\textwidth]{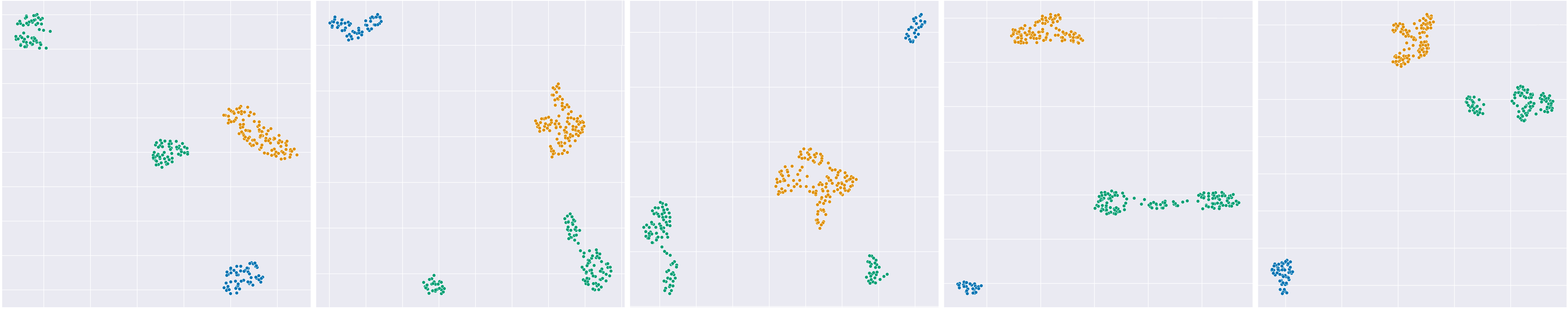}
        \caption*{UMAP Visualizations of $f^s_{cls}$ for AR images (Natural, Web, Graphics).}
        \label{fig:umap_ar}
    \end{minipage}
    
    \vspace{2mm} 
    
    \begin{minipage}{\textwidth}
        \centering
        \includegraphics[width=\textwidth]{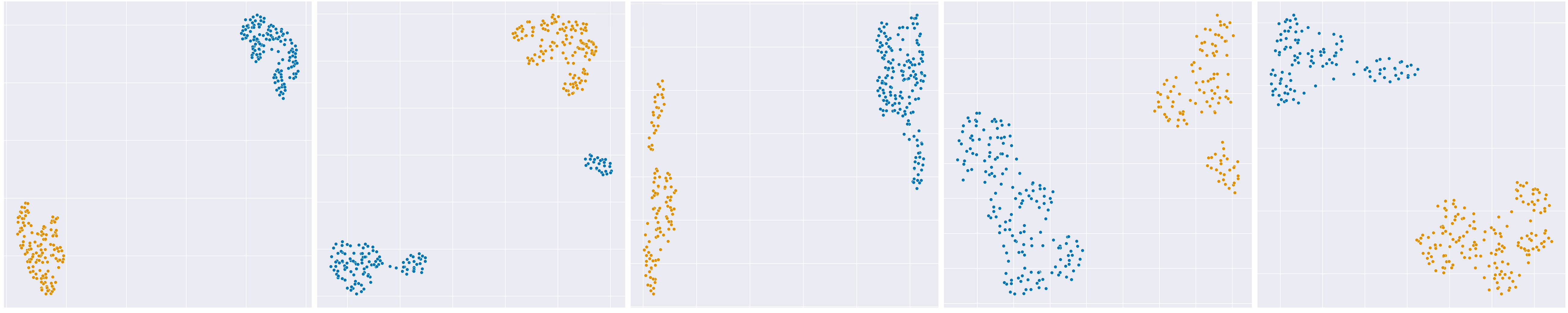}
        \caption*{UMAP Visualizations of $f^s_{cls}$ for Background images (Outdoor vs. Indoor) }
        \label{fig:umap_bg}
    \end{minipage}

    \caption{UMAP \cite{mcinnes2018umap} Visualizations of $f^s_{cls}$ of the AR images and the Background images across 5 folds. Each column represents a fold.}
    \label{fig:umap}
\end{figure*}

Based on our analysis, we observe notable distinctions in the attention mechanisms of TransformAR's encoders pre-trained on ImageNet in an unsupervised manner, and those fine-tuned on the ARIQA dataset with a category-specific feedback. When the encoders are fine-tuned, the $x_{\text{class}}$ token demonstrates an enhanced focus across different patches within the AR image, indicating increased learning diversity across heads. Moreover, background understanding leads to more nuanced attention distribution, mitigating biases towards brighter features like the sky in the outdoor scenarios. In the presence of visual confusion and distortion, the fine-tuned model maintains a consistent focus on the AR object across most heads, highlighting robustness against such challenges.


\subsection{UMAP visualization}
Figure \ref{fig:umap} shows the distribution of the reduced representations obtained by the superimposed encoder $\mathcal{F}^s(\cdot)$, specifically focusing on the $x_{\text{class}}$ representations for each class. This distribution is visualized using UMAP \cite{mcinnes2018umap}, a dimensionality reduction technique. The figure showcases how effectively the TransformAR-KD model can distinguish and segregate different classes within both the foreground and the background. Notably, the encoder $\mathcal{F}^s(\cdot)$ was not explicitly trained to predict classes; instead, its training focused on aligning the final representation $f^s_{cls}$ with ground-truth representations $\hat{f}^a_{cls}$ and $\hat{f}^b_{cls}$, as discussed in Section \ref{ssec:enhc_rep}. This alignment process enables the model to gather comprehensive knowledge about both foreground and background elements in the presence of visual confusion.

\section{Conclusion}
\label{sec:conclusion}
In this study, we introduced a novel approach for FR-IQA tailored for AR scenarios, namely TransformAR, along with its enhanced versions TransformAR-KD and TransformAR-KD+. Our methodology revolves around leveraging content-aware encoders to capture balanced low-level and high-level features. These features are then used to produce shift representations, effectively capturing distortions and visual confusion effects to generate final quality scores. Additionally, we enhance the encoders through knowledge distillation by maximizing the cosine similarity loss between the superimposed representations and the reference representations, further refining the model's performance. Addressing the challenge of data scarcity, we pursued model simplification by employing two blocks for the encoder and one block for the decoder. Furthermore, our model incorporates elastic net regularization and implements label smoothing to improve robustness against overfitting. Our comprehensive experimentation demonstrated that the proposed method not only improves upon existing results but also enhances attention maps. This validates our initial hypothesis, as outlined in the introduction, and underscores the ability of our model to provide foreground and background characteristics to the encoders, mirroring the natural observation of the humans.

Looking ahead, mitigating data scarcity and seeking more realistic AR scenarios, constructing new datasets specifically designed for AR-IQA remains a crucial aspect of future research. These datasets should account for the unique characteristics of AR technology, including factors such as binocular visual confusion, to facilitate the development of more effective objective quality assessment metrics.

\section*{Acknowledgments}
This work is partially funded by the Nouvelle-Aquitaine Research Council under project REALISME AAPR2022-2021-17027310.

This work has been submitted to the IEEE for possible publication. Copyright may be transferred without notice, after which this version may no longer be accessible.

\bibliographystyle{IEEEtran}
\bibliography{main.bib}

\end{document}